\documentclass{article}

\PassOptionsToPackage{numbers, compress}{natbib}
\usepackage[final]{nips_2017}

\usepackage[utf8]{inputenc} 
\usepackage[T1]{fontenc}    

\usepackage{amsfonts,amsmath,amssymb,amsthm}
\usepackage{graphicx}
\DeclareGraphicsExtensions{.pdf,.png,.gif,.jpg}

\theoremstyle{plain}

\theoremstyle{definition}

\theoremstyle{remark}

\providecommand{\delete}[1]{}
\providecommand{\nitro}[1]{}

\newcommand{\iid}{i.\@i.\@d.\@ }

\title{Adversarial Phenomenon in the Eyes\\ of Bayesian Deep Learning}

\author{
  Ambrish Rawat, Martin Wistuba, and Maria-Irina Nicolae \\
  IBM Research AI -- Ireland\\
  Mulhuddart, Dublin 15, Ireland\\
  \texttt{ambrish.rawat@ie.ibm.com, \{martin.wistuba,maria-irina.nicolae\}@ibm.com} \\
}

\begin{document}

\maketitle

\begin{abstract}

Deep Learning models are vulnerable to adversarial examples, i.e.\ images obtained via deliberate imperceptible perturbations, such that the model misclassifies them with high confidence.
However, class confidence by itself is an incomplete picture of uncertainty. We therefore use principled Bayesian methods to capture model uncertainty in prediction for observing adversarial misclassification.
We provide an extensive study with different Bayesian neural networks attacked in both white-box and black-box setups.
The behaviour of the networks for noise, attacks and clean test data is compared.
We observe that Bayesian neural networks are uncertain in their predictions for adversarial perturbations, a behaviour similar to the one observed for random Gaussian perturbations.
Thus, we conclude that Bayesian neural networks can be considered for detecting adversarial examples.

\end{abstract}

\section{Introduction}

Although massively successful, Deep Learning as conventionally practiced does not account for all forms of uncertainty.
For instance, when point estimates of parameters are learned, the desired uncertainty about parameters is not represented in the predictions.
These uncertainties can be useful doxastic indicators of a model's confidence in its own predictions.
Therefore, a careful examination of uncertainties must be a precursor to decision processes based on Deep Learning models.
This is extremely pertinent given that applications such as autonomous cars and medical diagnosis make large-scale use of Deep Learning models.
In these critical applications, it is essential that the used prediction models are reliable, and uncertainty in prediction provides one means to establish trust.

A more recent concern regarding reliability of Deep Learning models sources from their exhibited vulnerability to adversarial attacks and in particular adversarial images~\cite{biggio2013evasion,szegedy2013intriguing}.
These attacks correspond to images deliberately crafted by adding an imperceptible perturbation which is computed with an intent to obtain a high-confidence misclassification from a model.
The discovery of adversarial examples has challenged the reliability of Deep Learning models.
Unsurprisingly, the field of Adversarial Machine Learning has gained an overwhelming excitement among researchers, resulting in an arsenal of available methods for attacking \cite{Goodfellow2014,jsma,rndfgsm,carlini2017towards,moosavi2016deepfool} and defending \cite{zantedeschi2017efficient,labelsmoothing,feature1,szegedy2013intriguing} an image classification model.
In the light of these advances, it is imperative that uncertainty is accounted for before entrusting predictions obtained from these models \cite{gal2016uncertainty,kendall2017uncertaintie}.

Bayesian methods offer a principled way to represent uncertainties in a model and can therefore be utilised to quantify a model's confidence in its prediction.
Moreover, these models are inherently robust, as inference during both posterior-learning and prediction involves marginalisation over parameter uncertainties (in the form of prior and posterior respectively).
Following the elegant principles of probability theory, inference can be easily formulated for a Deep Learning model.
However, their massively parametrised non-linear nature makes Bayesian inference for these models analytically and computationally intractable.
Bayesian neural networks (BNNs) have been an element of extensive research for the last three decades including the seminal works of \cite{mackay1992practical}, \cite{neal2012bayesian}, and \cite{hinton1993keeping}.
However, practical approaches have only surfaced recently~\cite{graves2011practical,hernandez2015probabilistic,soudry2014expectation,blundell2015weight,Louizos2016,gal2016dropout} with efficient, scalable, and reliable methods still being an open challenge.
Some of these works belong to two broad classes of approximate inference - variational inference (VI) and assumed density filtering (ADF) ~\cite{opper1998bayesian} which we briefly describe below.

Variational inference is an approach which posits a family of approximate posteriors and proposes to choose the one that is closest to the exact posterior as measured by Kullback-Leibler divergence.
The optimisation assumes an equivalent form where a lower bound to the marginal likelihood is optimised for learning the parameters.
\cite{graves2011practical} and \cite{blundell2015weight} developed variational approaches for parameter inference.
As opposed to fully factored assumptions in these, \cite{Louizos2016} learns a matrix variate Gaussian posterior that accounts for correlations between weights.
The work of~\cite{gal2016dropout} grounded dropout as VI in a deep Gaussian process approximation of a neural network enabling a remarkably simple and highly practical way to propagate parameter uncertainty into a model's prediction.

Alternatively, assumed density filtering provides an online approach~\cite{opper1998bayesian} that incrementally updates the posterior over parameters for every new evidence.
Assuming a parametrised posterior distribution over the parameters, this approach iteratively incorporates each factor (prior and likelihood) by minimising the KL-divergence between the new approximation and the product of new factor and previous approximation.
\cite{soudry2014expectation} and \cite{hernandez2015probabilistic} developed learning algorithms based on expectation backpropagation and probabilistic backpropagation which scale this approach for neural networks.
\cite{ghosh2016assumed} further advanced their work, enabling multi-class classification.


As a first premise, one can theorise robustness to adversarial perturbation for BNNs given their capacity to represent and propagate different forms of uncertainties.
Robustness of Bayesian models to adversarial images has recently been explored in \cite{bradshaw2017adversarial}.
Similarly, \cite{LiGal2017Alpha} and \cite{Louizos2017} examine epistemic uncertainty in the form of predictive entropy for adversarial images.
However, BNNs are no exceptions to being vulnerable to adversarial attacks which leads to the second premise - a model albeit confident in its erroneous class prediction for an adversarial image will be potentially uncertain in the prediction itself.

The current work examines the adversarial phenomenon for BNNs through the lens of uncertainties in prediction.
Our study comprises four different Bayesian approaches and studies model uncertainty for each of them through metrics such as mutual information, predictive entropy and variation ratio.
We prove that these models exhibit increased uncertainty when under attack, laying the grounds for an adversarial detection system based on them.
To this end, we adapt the standard adversarial attack FGSM~\cite{Goodfellow2014} to the case of BNNs through Monte Carlo sampling.
The experimental study also compares adversarial attacks against Gaussian noise introduced in the inputs, confirming the similarity between the uncertainties exhibited by the model in both cases.


\section{Theoretical Motivation}\label{sec:theoretical_motivation}

In supervised learning, a Deep Learning model is parametrised by a set of weights $\mathcal{W}$ and is trained with labeled data $D_N = \{x_i,y_i\}_{i=1}^N$ (where, $x \in X, y \in Y$).
Bayesian inference for these models involves learning a posterior distribution over weights, $p(\mathcal{W}|D_N)$.
This can then be utilised for obtaining predictions for an unseen observation in the form of a posterior predictive distribution
\begin{equation}
p\left(y^\ast|x^\ast,D_N\right)=\int p\left(y^\ast|x^\ast,\mathcal{W}\right)\ p\left(\mathcal{W}|D_N\right)\ d\mathcal{W} = \mathbb{E}_{p\left(\mathcal{W}|D_N\right)}\left[p\left(y^\ast|x^\ast,\mathcal{W}\right)\right] \enspace.
\end{equation}
The likelihood distribution $p(y|x,\mathcal{W})$ is modeled as a Gaussian distribution for regression and a categorical distribution for classification.

Owing to the non-linearities in a Deep Learning model, the above integral is often analytically intractable.
To overcome this intractability, approximate schemes like VI and ADF can be incorporated into the modeling.
The central idea in these approaches is to learn an approximate tractable posterior, $q_{\theta}(\mathcal{W})$ as opposed to the exact one. Here, $\theta$ correspond to the set of variational parameters.
The resulting approximate predictive distribution $q(y^\ast|x^\ast,D_N)$ can then be computed as $\mathbb{E}_{q_{\theta}(\mathcal{W})}[p(y^\ast|x^\ast,\mathcal{W})]$.
It is important to note two key points here.
First, for reliable prediction it is beneficial to marginalise over the learned posterior, exact or otherwise, so that appropriate uncertainty about parameters is propagated into the prediction.
This marginalisation is often ignored in traditional Deep Learning, where point estimates of the parameters $\mathcal{W}_{\text{est}}$ are utilised for prediction $p(y^\ast|x^\ast,\mathcal{W}=\mathcal{W}_{\text{est}})$.
And second, computation of the posterior predictive distribution $\mathbb{E}_{p(\mathcal{W}|D_N)}[p(y^\ast|x^\ast,\mathcal{W})]$ can be utilised for summarising parameter uncertainty.
This is particularly interesting for classification where a model's confidence in its prediction is not readily available.
For instance, in the cases when $\mathbb{E}_{p(\mathcal{W}|D_N)}[p(y^\ast|x^\ast,\mathcal{W})]$ is computed as a Monte Carlo (MC) estimate, different summaries of uncertainty (including predicted entropy $\mathbb{H}\left[y^\ast|x^\ast,D_N\right]$~\cite{freeman1965elementary} and variation ratio~\cite{shannon2001mathematical}) can be computed by accounting for the statistical dispersion in the predicted MC samples.
As noted in \cite{gal2016uncertainty}, mutual information between parameter posterior and predictive distribution can reveal the model's confidence in its prediction $\mathbb{I}\left(y^\ast,\mathcal{W}\ |\ x^\ast,D_N\right)$.
This has been previously utilised in the context of active learning in the works of \cite{mackay1992information} and \cite{houlsby2011bayesian}.
We call this measure of uncertainty Model Uncertainty as measured by Mutual Information (MUMMI) which is defined as,

\begin{equation}
 \operatorname{MUMMI} : =\mathbb{H}\left[y^\ast|x^\ast,D_N\right] - \mathbb{E}_{p\left(\mathcal{W}\ |\ D_N\right)}\left[\mathbb{H}\left[y^\ast|x^\ast,\mathcal{W}\right]\right]\enspace.
\end{equation}

For the sake of completeness, we also repeat the definition of variation ratio and predicted entropy:
\begin{align}
\text{Predicted Entropy} &:= \mathbb{H}\left[y^\ast|x^\ast,D_N\right] \\
\text{Variation Ratio} &:=1-\frac{f}{M}\enspace,
\end{align}
where $f$ is the frequency of the mode.
Thus, the variation ratio is the relative number of times the majority class is not predicted.
It equals zero if the same class is predicted for all $M$ Monte Carlo samples.
The maximum $1-\left|Y\right|/M^{2}$ is reached when every class option is predicted equally.
Thus, the higher the variation ratio, the higher the confusion of the model.

\paragraph{Probabilistic Back Propagation (PBP)} 
\label{par:probabilistic_back_propagation_}
PBP \cite{hernandez2015probabilistic} makes several assumptions in its modeling approach.
First, it assumes an approximate posterior which is factored across all model parameters and where each factor takes the form of a Gaussian $q_\theta(\mathcal{W}) =\prod_{w\in \mathcal{W}}\mathcal{N}(w;m_w,v_w)$.
Thus, the PBP model consists of twice the number of parameters then a conventional neural network.
In order to learn these parameters, PBP sequentially incorporates each of the $N$ likelihood terms $l(\mathcal{W})$ by minimising the KL-divergence between $s(\mathcal{W})=Z^{-1}l(\mathcal{W})q_{\theta_{\text{old}}}(\mathcal{W})$ and $q_{\theta_{\text{new}}}(\mathcal{W})$.
This optimisation is performed with respect to parameters of the approximate posterior, which for the Gaussian case yields elegant update rules for posterior parameters that depend solely on derivatives of $\log Z$ \cite{Minka2001}.
Further, for simplifying the analytic computation of $Z$, PBP assumes that the distribution of the output of the neural network is a multivariate Gaussian.
The computation of $Z$ then amounts to a forward pass in the network where the first and second moments of outputs from one layer are fed into its following layer, eventually yielding the parameters of the aforementioned multivariate Gaussian. The expressions for this moment propagation for ReLU activation functions are mentioned in \cite{hernandez2015probabilistic}.
Neural networks for classification utilise a softmax operation over the output obtained from neural network to yield probability vectors over classes yielding the likelihood and subsequently the required $Z$.
\cite{ghosh2016assumed} employ the reparametrisation trick~\cite{Kingma2015} to compute the derivatives of this $Z$.

It is worth noting that, with the multivariate Gaussian assumption, the computation of $E_{q_\theta(\mathcal{W})}[p(y^\ast|x^\ast,\mathcal{W})]$ gets simplified for PBP.
With a single forward pass for a data point $x^\ast$, one can obtain the parameters of the Gaussian distribution corresponding to the output of the network.
Thus, instead of sampling from the approximate posterior, one can equivalently sample from the output distribution.
The expressions for this equivalent marginalisation are described in the appendix.


\paragraph{Variational Matrix Gaussian (VMG)}
\label{par:vmg}
\cite{Louizos2016} introduce a variational BNN that treats the weight matrix as a whole using matrix variate Gaussian distributions~\cite{Gupta1999}.
This modeling reduces the number of variance-related parameters to estimate and introduces row- and columnwise correlations.
For an efficient computation, the full covariance matrix is replaced with a diagonal approximation, rendering posterior uncertainty estimation easier through information sharing.
The local reparametrisation trick~\cite{Kingma2015} exposes the deep Gaussian process nature of the model, allowing for more efficient sampling through the use of pseudo-data.

\paragraph{Bayes by Backprob (BBB)}
\label{par:bbb}
Bayes by Backprob \cite{blundell2015weight} is a variational approach to Bayesian neural networks.
The variational posterior is assumed to be a diagonal Gaussian distribution.
This assumes independence between variables and consequently only twice as many parameters are needed in comparison to a standard multilayer perceptron (MLP).
The authors propose the use of a scale mixture of two Gaussian densities as a prior:
\begin{equation}
 p\left(\mathcal{W}\right)=\prod_{j}\alpha\mathcal{N}\left(\mathcal{W}_{j}\ |\ 0,\sigma_{1}^{2}\right)+\left(1-\alpha\right)\mathcal{N}\left(\mathcal{W}_{j}\ |\ 0,\sigma_{2}^{2}\right)\enspace.
\end{equation}
Both Gaussian distributions have zero mean, but can have differing variances.
The hyperparameter $\alpha$ weighs the influence of the densities.
The authors claim better results using the scale mixture of Gaussians in comparison to a Gaussian prior, i.e.\ $\alpha=0$ or $\alpha=1$, respectively.

\paragraph{MC Dropout} 
The work of \cite{gal2016dropout} establishes dropout as approximate variational inference in a Gaussian Process interpretation of neural networks.
\cite{Gal2016Bayesian} build on this further and use a specific Bernoulli approximating variational distributions to develop approximate variational inference in BNNs, which they define as MC-Dropout.
Further, this allows them to impart a Bayesian interpretation to convolutional operations making MC-Dropout a highly scalable method.
Additionally, \cite{Gal2016Bayesian} observe that MC-Dropout with dropouts used in both convolutional and fully connected layers has the best performance gain w.r.t.\ its corresponding standard dropout-trained model.
\label{par:mc_dropout}


%

\section{Attacking Bayesian Neural Networks}
\label{sec:attacking_bnns_with_fgsm}
For a standard neural network trained for multi-class classification and a data point $x$ of label $y$, an adversarial attack aims to find an adversarial perturbation $\Delta x$ such that the modified input $x + \Delta x$ is not assigned to class $y$ anymore\footnote{When the true label is unavailable, the model's prediction can be used as a label $y_{\text{pred}}$. The attack becomes untargeted, aiming to change the predicted class.}.
The fast gradient sign method (FGSM)~\cite{Goodfellow2014} proposes to compute the noise to be added as a small mass of the gradient of the loss function:
\begin{equation} \label{eq:fgsm}
  \Delta x = \epsilon \cdot \text{sign}\nabla_x J(p(y|x,\mathcal{W}=\mathcal{W_{\text{est}}}),y)\enspace,
\end{equation}
where $\epsilon > 0$ is the magnitude of the attack, $J$ is the standard loss function for classification (negative log-likelihood, also known as categorical cross-entropy, between the predicted probability and true label), and $\mathcal{W_{\text{est}}}$ represent the point estimates of weights learned during training.

In the case of Bayesian neural networks, the aim is to learn the approximate posterior which in part can involve the maximisation of a likelihood term, but this maximisation is often not at the focal point of the optimisation objective.
For instance, while the expected log-likelihood features as a key term in the variational objective, the case is harder to establish for methods like expectation propagation.
However, both standard and Bayesian neural networks aim to achieve low misclassification rate; we thus use negative log-likelihood as loss function when crafting adversarial samples against BNNs.
In their case, the predicted probability takes the form of $\mathbb{E}_{q_{\theta}(\mathcal{W})}[p(y^\ast|x^\ast,\mathcal{W})]$, value that can be estimated with $M$ Monte Carlo samples.
Consequently, we use the following approximation of the gradient of the loss function:
\begin{equation}
 \nabla_x J\left(\mathbb{E}_{q_{\theta}\left(\mathcal{W}\right)}\left[p\left(y|x,\mathcal{W}\right)\right],y\right) \approx \nabla_{x}J\left(\frac{1}{M}\sum_{i=1}^{M}p\left(y|x,\mathcal{W}^{\left(i\right)}\right),y\right)\enspace , \label{eq:approx1}
\end{equation}
where $\mathcal{W}^{\left(i\right)}$ is the i-th Monte Carlo sample.

Since we are using the negative log-likelihood as our loss function, we reformulate the gradient in order to efficiently compute it in current neural network libraries.
Without loss of generality, we assume that there is only one true class $y_{l}$:
\begin{equation}
 \nabla_{x}J\left(\frac{1}{M}\sum_{i=1}^{M}p\left(y|x,\mathcal{W}^{\left(i\right)}\right),y\right)=\frac{\sum_{i=1}^{M}p\left(y=y_{l}|x,\mathcal{W}^{\left(i\right)}\right)\nabla_{x}J\left(p\left(y|x,\mathcal{W}^{\left(i\right)}\right),y_{l}\right)}{\sum_{i=1}^{M}p\left(y=y_{l}|x,\mathcal{W}^{\left(i\right)}\right)}\enspace .
\end{equation}
Detailed derivations can be found in the Appendix. Alternatively, one can use a perturbation computed as $\mathbb{E}_{q_{\theta}(\mathcal{W})}[\nabla_x J\left(p(y|x,\mathcal{W}),y\right)]$. We find the behaviour of uncertainties for this formulation of attack to be similar to the one from Equation~\eqref{eq:approx1}.





\section{Experiments and Discussion}\label{sec:Conclusions}

We evaluate and contrast three different estimates of uncertainty - predicted entropy, variation ratio, and MUMMI for four different Bayesian neural network models - Bayes by Backprob (BBB)~\cite{blundell2015weight}, Probabilistic Backpropagation (PBP)~\cite{ghosh2016assumed}, Variational Matrix Gaussian (VMG)~\cite{Louizos2016} and MC-Dropout~\cite{gal2016dropout}.
All of these models are trained for classification on the famous MNIST dataset~\cite{Lecun1998} with training parameters adopted from the respective works. The training setups for the models are specified in the appendix.

Our experiments pivot on the central idea of analysing the extrapolation behaviour of prediction models, i.e.\ prediction uncertainties at points that lie away from the training data.
In classification and regression tasks, a model predicts confidently for a data sample that is similar to training samples, and underconfidently when queried with an example that is dissimilar and far way from the data.
The further such an example is, the more underconfident the prediction will be.
This behaviour is characteristic to probabilistic models like Gaussian processes.
In regression, the low confidence is reflected in high variance of the predictive distribution at the queried sample.
However, for classification one needs to inspect both the class confidence and the model uncertainty in prediction to observe this behaviour\footnote{It can be argued that the softmax operation on the outputs of a network could result in misleading class confidence probabilities.
A diagnosis in terms of variance of the network output vectors prior to softmax can be used to analysed to contest such a claim.
However, we do not pursue this in the scope of current work.}.
We study this across three different setups.
In the first setup, we inspect uncertainties in the direction of adversarial perturbations.
We then contrast this with random Gaussian noise.
Finally, we take a closer look at uncertainties by visualising their distribution for different image sets.
This provides useful insights into the workings of different models.

\subsection{Adversarial Perturbation and Model Uncertainty} 
\label{par:adversarial_perturbation_and_model_uncertainty}

\begin{figure}[ht]
    \centering
    \includegraphics[width=\linewidth]{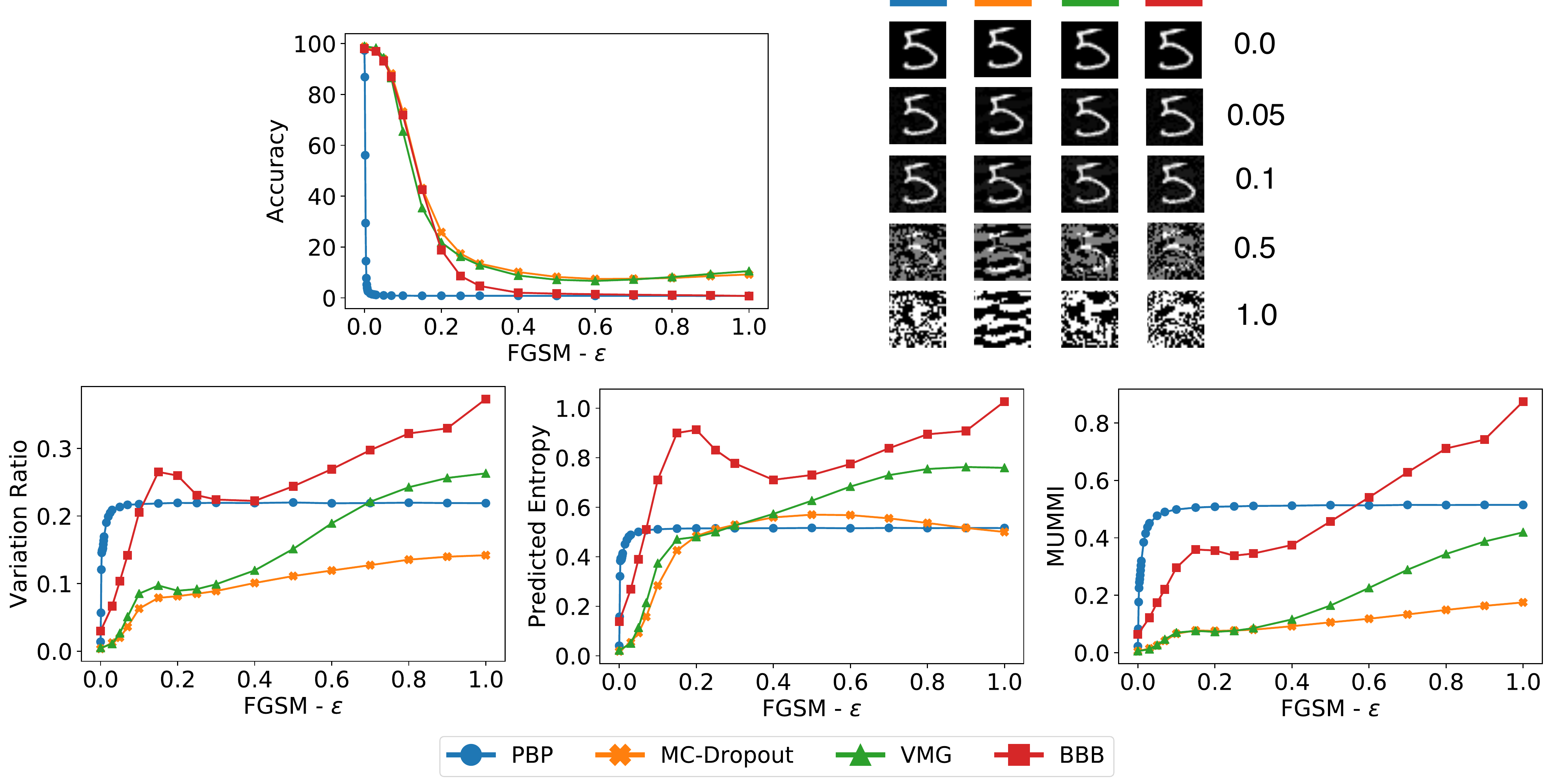}
    \caption{Top left plot shows a rapid decrease in accuracy for increasing attack strength $\varepsilon$.
    Top right shows exemplary adversarial images for each of the networks.
    The bottom column shows a significant increase of uncertainty with respect to three metrics when under attack.
    A larger range of epsilons is used for PBP to capture the behaviour for very small values.}
    \label{fig:whitebox}
\end{figure}

For this experiment, we craft adversarial images for the four models in a white-box setup using the modified FGSM proposed in Equation \eqref{eq:approx1}.
The perturbations are generated for the MNIST test set.
A drop in accuracy for imperceptible changes is evident in Figure \ref{fig:whitebox}.
In theory, propagating appropriate uncertainties into class predictions can impart the necessary robustness to adversarial attacks.
However, BNNs are not found to be robust to adversarial attacks, as has been observed in ~\cite{bradshaw2017adversarial,LiGal2017Alpha}.

An alternate advantage point of model's confidence in its own prediction is available in the form of predictive entropy, variation ratio and MUMMI.
Each of these summaries of uncertainty are observed to have significantly high values for adversarial images in comparison to the ones obtained for the MNIST test set.
This clearly indicates that the although confident in class prediction, BNN models are uncertain in their own prediction.
Arguably, the models \textit{know what they don't {}know, about what they don't know}.

\subsection{Gaussian Perturbation and Model Uncertainty} 
\label{par:gaussian_perturbation_and_model_uncertainty}

\begin{figure}[ht]
    \centering
    \includegraphics[width=\linewidth]{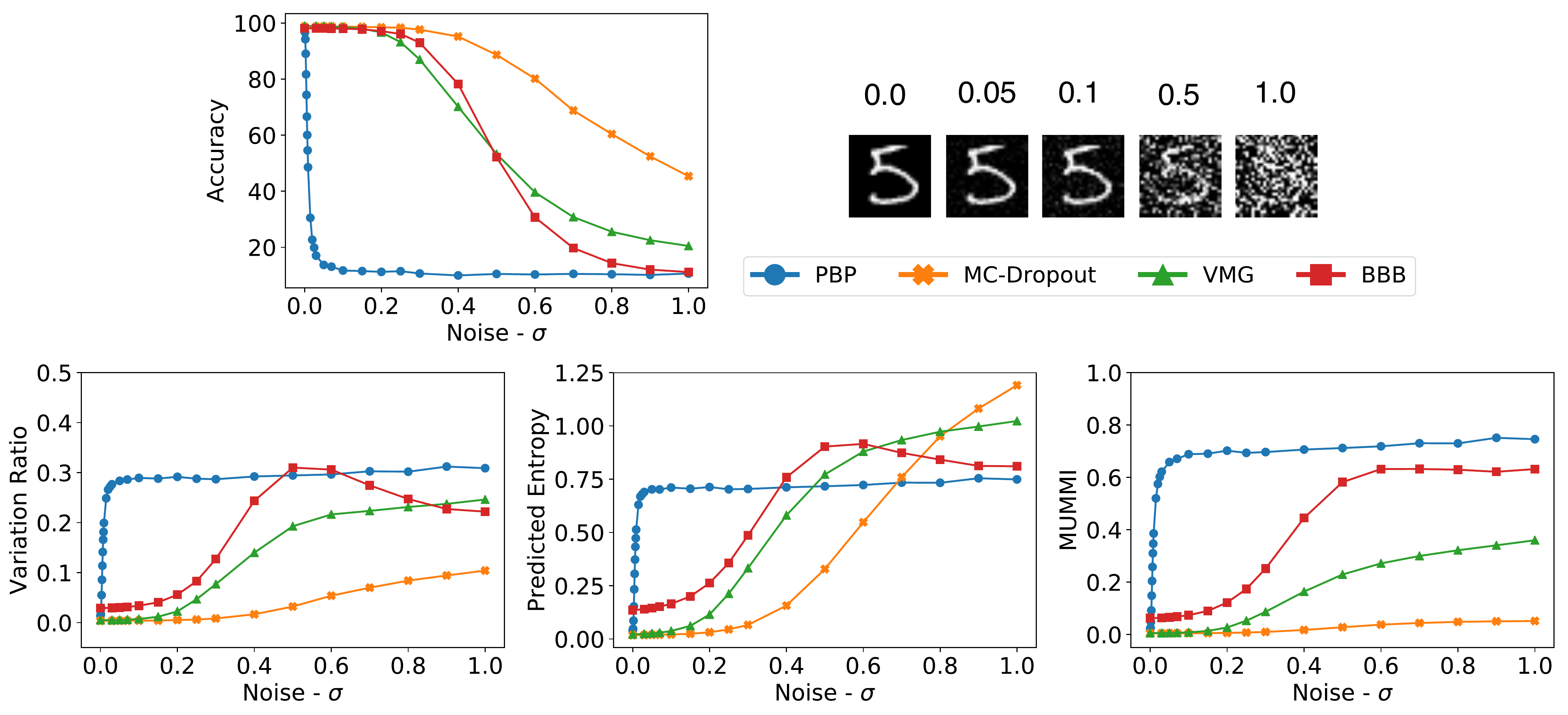}
    \caption{Similar results as presented in Figure \ref{fig:whitebox}. However, since the noise is not adversarial, the accuracy does not drop as fast for most methods.}
    \label{fig:mnist-pertb}
\end{figure}

For this experiment, we perturb the MNIST test set with Gaussian noise of zero mean and varying standard deviation.
As shown in Figure \ref{fig:mnist-pertb}, the accuracy drops gradually with perturbation increase, with a significant drop corresponding to perturbations $\sigma\geq 0.5$ that lead to perceptible and visually detectable distortions.
Similarly, the model uncertainty in its own prediction increases with perturbation strength.
Thus, the further a sample is from the training data distribution, both the class prediction and model confidence in prediction are low.
Adversarial examples present an intriguing case, where the rapid drop in accuracy with small imperceptible changes to images suggests that class confidence by itself might be an incomplete picture of model confidence.
Interestingly, the uncertainties demonstrate a very similar behaviour to adversarial perturbations.
This indicates that BNNs could be perceiving adversarial images as samples that lie away from training set distribution.

\subsection{Uncertainty Footprints} 
\label{par:uncertainty_footprints}

For a closer surgical analysis, we visualise different sets of points from the data space along two different uncertainty axes, predicted class probability and model uncertainty in its prediction.
For brevity we only include the plots corresponding to MUMMI in the main paper.
Every model leaves a unique signature corresponding to distribution of these uncertainties for different sets.
We call these plots uncertainty footprints.
We analyse the quality of these footprints at eight different regions in the data space:
\paragraph{MNIST test set}
This serves as the idealised region that is closest to the training samples, where we expect the model to have high class confidence, as well as high certainty in its own prediction.
\paragraph{Adversarial images}
This corresponds to the specific regions where Deep Learning models are known to exhibit high class confidence in a wrong class.
However, we expect to observe a model to be highly underconfident in its own prediction.
For a fair comparison, we generate these images on an independently trained CNN-classifier for MNIST using the standard FGSM attack.
\paragraph{Perturbed MNIST test set}
We create samples by perturbing clean images with a random Gaussian noise of varying strength $\sigma\cdot\mathcal{N}(\boldsymbol{0},\boldsymbol{I})$.
\paragraph{Noise samples}
Intuitively, high underconfidence in its own prediction is characteristic for noisy regions in the data space which correspond to no meaningful image.
We create three different synthetic noise sets for this experiment
\begin{itemize}
 \item Uniform: each pixel is independently sampled from a uniform distribution.
 \item Pixel: each pixel is independently sampled from a Gaussian distribution whose moments are estimated from the MNIST training set distribution.
 \item MVN: pixels are jointly sampled from a multivariate normal distribution with mean and covariance estimates obtained from the MNIST training set.
\end{itemize}

\begin{figure}[!htb]
    \centering
    \includegraphics[width=\linewidth]{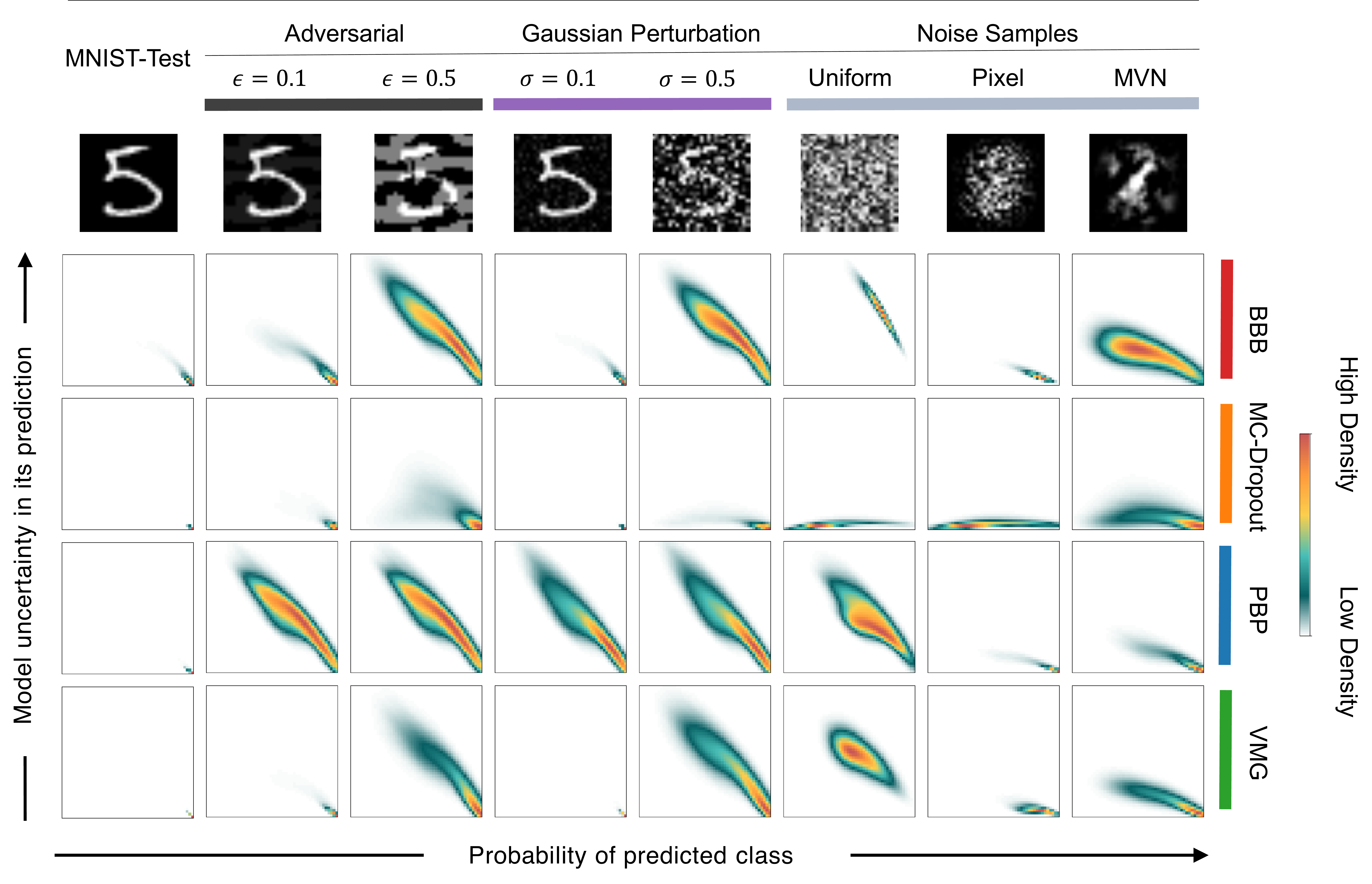}
    \caption{Scatter plot of dataset points along two summaries of uncertainty - i) probability of predicted class (x-axis) and MUMMI (y-axis), overlayed with the density of points.
    A representative sample from each of the set is included for clarity.}
    \label{fig:mutual-inf}
\end{figure}

Figure \ref{fig:mutual-inf} brings out some interesting features of these different models. A dense collection of points in bottm-right corners for MNIST Test set (leftmost column) indicates both high-certainty in class prediction and high-confidence in the prediction itself. The evident similarity of the footprints of the two sets - Adversarial ($\epsilon=0.5$) and Gaussian Perturbation $(\sigma=0.5)$ clearly indicates that BNNs recognise adversarial pertubations as \textit{moving away from data distribution}. Footprints on Uniform noise present an idealised scenario where both class-confidence and confidence in prediction are low. It is worth noting that PBP admits a sharp rise in uncertainty summaries for both adversarial and Gaussian perturbations (also noted in Figure \ref{fig:whitebox} and \ref{fig:mnist-pertb}). Interestingly, the behaviour is consistent for very low undetectable perturbations. However, it displays a starking high confidence on uninterpretable Pixel and MVN noises. Other models also have varying and unintuitive behaviour on predictions for these two noises. For instance, the high class confidence of in the predictions of MC Dropout for MVN set can be explained by its convolutional architechture. The models incorporate different preprocessing techniques including feature normalisation and data normalisation which can potentially reflect in the observed behaviour on noise sets.


\section{Conclusion}

Bayesian Neural Networks are vulnerable to adversarial attacks. In this work, we analysed different forms of uncertainty for adversarial images as obtained from different Bayesian Neural Networks.
We appropriately modified the standard form of the FGSM attack for the case of BNNs and studied model uncertainty for the images thus obtained.
We summarised model uncertainty in three different measures - MUMMI, predictive entropy, and variation ratio.
There is clear evidence of increasing model uncertainty with attack strength which indicates that these quantifications can be used to build effective adversarial detection systems.
We seek to develop this further in our future work.

We contrasted the study of adversarial perturbation against random Gaussian perturbation and observed that the quality of model uncertainty on the two sets is very similar.
This substantiates a known facet of probabilistic modeling where a model predicts underconfidently for points that are highly dissimilar from the data distribution.

Owing to the different approximations incorporated in the training of Bayesian Neural Networks, the quality of prediction uncertainties varies across models.
There are two important aspects to this. First, the probability predictions from the model may not be well-calibrated~\cite{guo2017calibration}.
Second, while Deep learning models are designed for representation learning, Bayesian modeling has its emphasis on uncertainty representation and propagation.
Thus, it is difficult to argue about the quality of learnt representations for BNNs. This is also reflected in the uncertainty footprints obtained for noise sets.
Although speculative, we wish to formally investigate the effect of Bayesian methods on representation learning in neural network as future research.

Most of the used BNNs in this study utilise MLP architectures which do not scale for high dimensional coloured images, with the exception of MC-Dropout.
Therefore, a detailed study for MC-Dropout is required to demonstrate the applicability of model uncertainty for adversarial detection on various state-of-the-art attacks. Additionally, we believe that transferability of attacks between different BNNs can help understand the usefulness of each different part of the prediction pipeline. For instance, detection and classification can be separated into two different tasks and consequently be handled by two different models.

\bibliographystyle{plain}
\bibliography{sample}

\begin{thebibliography}{10}

\bibitem{biggio2013evasion}
Battista Biggio, Igino Corona, Davide Maiorca, Blaine Nelson, Nedim
  {\v{S}}rndi{\'c}, Pavel Laskov, Giorgio Giacinto, and Fabio Roli.
\newblock Evasion attacks against machine learning at test time.
\newblock In {\em Joint European Conference on Machine Learning and Knowledge
  Discovery in Databases}, pages 387--402. Springer, 2013.

\bibitem{blundell2015weight}
Charles Blundell, Julien Cornebise, Koray Kavukcuoglu, and Daan Wierstra.
\newblock Weight uncertainty in neural network.
\newblock In {\em International Conference on Machine Learning}, pages
  1613--1622, 2015.

\bibitem{bradshaw2017adversarial}
John Bradshaw, Alexander G de~G Matthews, and Zoubin Ghahramani.
\newblock Adversarial examples, uncertainty, and transfer testing robustness in
  gaussian process hybrid deep networks.
\newblock {\em arXiv preprint arXiv:1707.02476}, 2017.

\bibitem{carlini2017towards}
Nicholas Carlini and David Wagner.
\newblock Towards evaluating the robustness of neural networks.
\newblock In {\em Security and Privacy (SP), 2017 IEEE Symposium on}, pages
  39--57. IEEE, 2017.

\bibitem{freeman1965elementary}
Linton~C Freeman.
\newblock {\em Elementary applied statistics: for students in behavioral
  science}.
\newblock John Wiley \& Sons, 1965.

\bibitem{gal2016uncertainty}
Yarin Gal.
\newblock {\em Uncertainty in deep learning}.
\newblock PhD thesis, University of Cambridge, 2016.

\bibitem{Gal2016Bayesian}
Yarin Gal and Zoubin Ghahramani.
\newblock Bayesian convolutional neural networks with {B}ernoulli approximate
  variational inference.
\newblock In {\em 4th International Conference on Learning Representations
  (ICLR) workshop track}, 2016.

\bibitem{gal2016dropout}
Yarin Gal and Zoubin Ghahramani.
\newblock Dropout as a bayesian approximation: Representing model uncertainty
  in deep learning.
\newblock In {\em International Conference on Machine Learning (ICML)}, pages
  1050--1059, 2016.

\bibitem{ghosh2016assumed}
Soumya Ghosh, Francesco~Maria Delle~Fave, and Jonathan~S Yedidia.
\newblock Assumed density filtering methods for learning bayesian neural
  networks.
\newblock In {\em AAAI}, pages 1589--1595, 2016.

\bibitem{Goodfellow2014}
Ian~J. Goodfellow, Jonathon Shlens, and Christian Szegedy.
\newblock Explaining and harnessing adversarial examples.
\newblock {\em CoRR}, abs/1412.6572, 2014.

\bibitem{graves2011practical}
Alex Graves.
\newblock Practical variational inference for neural networks.
\newblock In {\em Advances in Neural Information Processing Systems (NIPS)},
  pages 2348--2356, 2011.

\bibitem{guo2017calibration}
Chuan Guo, Geoff Pleiss, Yu~Sun, and Kilian~Q Weinberger.
\newblock On calibration of modern neural networks.
\newblock In {\em Proceedings of the 34th International Conference on Machine
  Learning (ICML-17)}, 2017.

\bibitem{Gupta1999}
Arjun~K. Gupta and Daya~K. Nagar.
\newblock {\em Matrix Variate Distributions}.
\newblock Monographs and Surveys in Pure and Applied Mathematics. Taylor \&
  Francis, 1999.

\bibitem{hernandez2015probabilistic}
Jos{\'e}~Miguel Hern{\'a}ndez-Lobato and Ryan Adams.
\newblock Probabilistic backpropagation for scalable learning of bayesian
  neural networks.
\newblock In {\em International Conference on Machine Learning}, pages
  1861--1869, 2015.

\bibitem{hinton1993keeping}
Geoffrey~E. Hinton and Drew Van~Camp.
\newblock Keeping the neural networks simple by minimizing the description
  length of the weights.
\newblock In {\em Conference on Computational Learning Theory (COLT)}, pages
  5--13. ACM, 1993.

\bibitem{houlsby2011bayesian}
Neil Houlsby, Ferenc Husz{\'a}r, Zoubin Ghahramani, and M{\'a}t{\'e} Lengyel.
\newblock Bayesian active learning for classification and preference learning.
\newblock {\em arXiv preprint arXiv:1112.5745}, 2011.

\bibitem{kendall2017uncertaintie}
Alex Kendall and Yarin Gal.
\newblock {What Uncertainties Do We Need in Bayesian Deep Learning for Computer
  Vision?}
\newblock In {\em Advances in Neural Information Processing Systems 30 (NIPS)},
  2017.

\bibitem{Kingma2015}
Diederik~P Kingma, Tim Salimans, and Max Welling.
\newblock Variational dropout and the local reparameterization trick.
\newblock In C.~Cortes, N.~D. Lawrence, D.~D. Lee, M.~Sugiyama, and R.~Garnett,
  editors, {\em Advances in Neural Information Processing Systems (NIPS)},
  pages 2575--2583. Curran Associates, Inc., 2015.

\bibitem{Lecun1998}
Yann Lecun, Léon Bottou, Yoshua Bengio, and Patrick Haffner.
\newblock Gradient-based learning applied to document recognition.
\newblock In {\em Proceedings of the IEEE}, pages 2278--2324, 1998.

\bibitem{LiGal2017Alpha}
Yingzhen Li and Yarin Gal.
\newblock Dropout inference in bayesian neural networks with alpha-divergences.
\newblock In {\em Proceedings of the 34th International Conference on Machine
  Learning (ICML-17)}, 2017.

\bibitem{Louizos2016}
Christos Louizos and Max Welling.
\newblock Structured and efficient variational deep learning with matrix
  gaussian posteriors.
\newblock In {\em International Conference on Machine Learning (ICML)}, pages
  1708--1716, 2016.

\bibitem{Louizos2017}
Christos Louizos and Max Welling.
\newblock Multiplicative normalizing flows for variational bayesian neural
  networks.
\newblock In {\em International Conference on Machine Learning (ICML)}, pages
  2218--2227, 2017.

\bibitem{mackay1992information}
David~JC MacKay.
\newblock Information-based objective functions for active data selection.
\newblock {\em Neural computation}, 4(4):590--604, 1992.

\bibitem{mackay1992practical}
David~JC MacKay.
\newblock A practical bayesian framework for backpropagation networks.
\newblock {\em Neural computation}, 4(3):448--472, 1992.

\bibitem{Minka2001}
Thomas~P. Minka.
\newblock {\em A Family of Algorithms for Approximate Bayesian Inference}.
\newblock PhD thesis, Cambridge, MA, USA, 2001.
\newblock AAI0803033.

\bibitem{moosavi2016deepfool}
Seyed-Mohsen Moosavi-Dezfooli, Alhussein Fawzi, and Pascal Frossard.
\newblock Deepfool: a simple and accurate method to fool deep neural networks.
\newblock In {\em IEEE Conference on Computer Vision and Pattern Recognition
  (CVPR)}, pages 2574--2582, 2016.

\bibitem{neal2012bayesian}
Radford~M Neal.
\newblock {\em Bayesian learning for neural networks}, volume 118.
\newblock Springer Science \& Business Media, 2012.

\bibitem{opper1998bayesian}
Manfred Opper and Ole Winther.
\newblock A bayesian approach to on-line learning.
\newblock {\em On-line Learning in Neural Networks, ed. D. Saad}, pages
  363--378, 1998.

\bibitem{jsma}
Nicolas Papernot, Patrick McDaniel, Somesh Jha, Matt Fredrikson, Z~Berkay
  Celik, and Ananthram Swami.
\newblock The limitations of deep learning in adversarial settings.
\newblock In {\em Security and Privacy (EuroS\&P), 2016 IEEE European Symposium
  on}, pages 372--387. IEEE, 2016.

\bibitem{shannon2001mathematical}
Claude~E Shannon.
\newblock A mathematical theory of communication.
\newblock {\em ACM SIGMOBILE Mobile Computing and Communications Review},
  5(1):3--55, 2001.

\bibitem{soudry2014expectation}
Daniel Soudry, Itay Hubara, and Ron Meir.
\newblock Expectation backpropagation: Parameter-free training of multilayer
  neural networks with continuous or discrete weights.
\newblock In {\em Advances in Neural Information Processing Systems (NIPS)},
  pages 963--971, 2014.

\bibitem{szegedy2013intriguing}
Christian Szegedy, Wojciech Zaremba, Ilya Sutskever, Joan Bruna, Dumitru Erhan,
  Ian Goodfellow, and Rob Fergus.
\newblock Intriguing properties of neural networks.
\newblock {\em arXiv preprint arXiv:1312.6199}, 2013.

\bibitem{rndfgsm}
Florian Tram{\`e}r, Alexey Kurakin, Nicolas Papernot, Dan Boneh, and Patrick
  McDaniel.
\newblock Ensemble adversarial training: Attacks and defenses.
\newblock {\em arXiv preprint arXiv:1705.07204}, 2017.

\bibitem{labelsmoothing}
David Warde-Farley and Ian Goodfellow.
\newblock 11 adversarial perturbations of deep neural networks.
\newblock {\em Perturbations, Optimization, and Statistics}, page 311, 2016.

\bibitem{feature1}
Weilin Xu, David Evans, and Yanjun Qi.
\newblock Feature squeezing: Detecting adversarial examples in deep neural
  networks.
\newblock {\em arXiv preprint arXiv:1704.01155}, 2017.

\bibitem{zantedeschi2017efficient}
Valentina Zantedeschi, Maria-Irina Nicolae, and Ambrish Rawat.
\newblock Efficient defenses against adversarial attacks.
\newblock In {\em ACM Workshop on Artificial Intelligence and Security
  (AISec)}, 2017.

\end{thebibliography}

\newpage
\appendix
\section{Appendix}

\subsection{Derivation for gradient computation}

We provide a derivation to show that Equation~\eqref{eq:approx1} holds.
By definition of the loss function $J$
\begin{equation}
\nabla_{x}J\left(\frac{1}{M}\sum_{i=1}^{M}p(y|x,\mathcal{W}^{\left(i\right)}),y\right)=\nabla_{x}\sum_{y_{l}\in Y}\left(-y_{l}\log\left(\frac{1}{M}\sum_{i=1}^{M}p\left(y=y_{l}|x,\mathcal{W}^{\left(i\right)}\right)\right)\right)\enspace.
\end{equation}

Deriving for $x$ yields
\begin{equation}
\sum_{y_{l}\in Y}\frac{\sum_{i=1}^{M}\nabla_{x}-y_{l}\ p\left(y=y_{l}|x,\mathcal{W}^{\left(i\right)}\right)}{\sum_{i=1}^{M}p\left(y=y_{l}|x,\mathcal{W}^{\left(i\right)}\right)}=\sum_{y_{l}\in Y}\frac{\sum_{i=1}^{M}p\left(y=y_{l}|x,\mathcal{W}^{\left(i\right)}\right)\frac{\nabla_{x}\left(-y_{l}\ p\left(y=y_{l}|x,\mathcal{W}^{\left(i\right)}\right)\right)}{p\left(y=y_{l}|x,\mathcal{W}^{\left(i\right)}\right)}}{\sum_{i=1}^{M}p\left(y=y_{l}|x,\mathcal{W}^{\left(i\right)}\right)}\enspace,
\end{equation}
which is by the definition of the gradient of the loss function
\begin{equation}
\sum_{y_{l}\in Y}\frac{\sum_{i=1}^{M}p\left(y=y_{l}|x,\mathcal{W}^{\left(i\right)}\right)\nabla_{x}J\left(p\left(y|x,\mathcal{W}^{\left(i\right)}\right),y_{l}\right)}{\sum_{i=1}^{M}p\left(y=y_{l}|x,\mathcal{W}^{\left(i\right)}\right)}\enspace.
\end{equation}

\subsection{Equivalent Marginalisations in PBP}

The modeling in PBP allows for moment propagation of input and output vectors for each layer of the network \cite{hernandez2015probabilistic}.
Under assumptions of PBP, the output vector $z_L$ in a $L$-layered network has a Gaussian distribution $\mathcal{N}(m^{z_L},v^{z_L})$, given the approximate posterior of network parameters.
In classification setting, softmax operation $\sigma$, is applied on the output vector $z^L$.
Thus the approximated predictive probability $\mathbb{E}_{q_\theta(\mathcal{W})}[p(y^\ast|x^\ast,\mathcal{W})]$ can be computed as $\mathbb{E}_{\mathcal{N}(m^{z_L},v^{z_L})}[\sigma(z_L)]$.
We use Monte Carlo estimates of this latter formulation in all our experiments (as also suggested in \cite{ghosh2016assumed}).

\subsection{Implementation Details}

For PBP and VMG, we train a MLP with 2 hidden layers of 400 units each. Similarly, for BBB we train a MLP with 2 hidden layers of 1200 units each.
For MC-Dropout we train a LeNet like CNN with 2 convolution layers and 1 dense layer of 500 units.
A dropout rate of 0.5 was used for the parameters of the dense layer while training.
During generation of adversarial images, the pixel values are clipped between the range $(0.0,1.0)$. We use $M=100$ samples for obtaining the predicted class probabilities from each of the BNNs. Similarly, for the gradient computation in FGSM, we used $M=100$ samples (Equation~\eqref{eq:approx1}).
\subsection{Black-Box Attacks from Standard MLP and Standard CNN}

\begin{figure}[!htb]
    \centering
    \includegraphics[width=\linewidth]{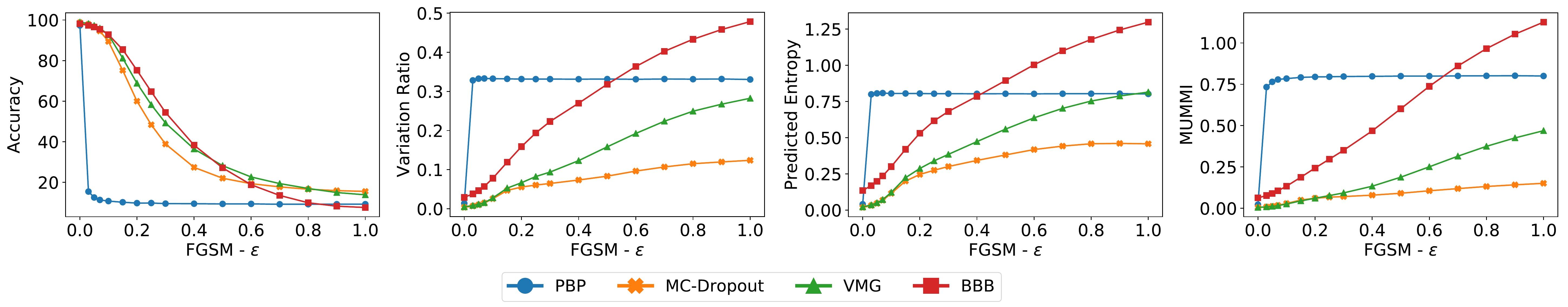}
    \caption{Accuracy and model uncertainties for adversarial images generated in a black-box setting on an independently trained CNN.}
    \label{fig:blackbox}
\end{figure}

We find that adversarial attacks easily transfer from standard neural networks to BNNs.
However, as with white-box setting, model uncertainty is observed to increase with attack strength (Figure~\ref{fig:blackbox}).
We craft adversarial images using FGSM on an independently trained CNN and obtain predictions from BNNs for the same.

\subsection{Training Set Distance}

\begin{figure}[!htb]
    \centering
    \includegraphics[width=\linewidth]{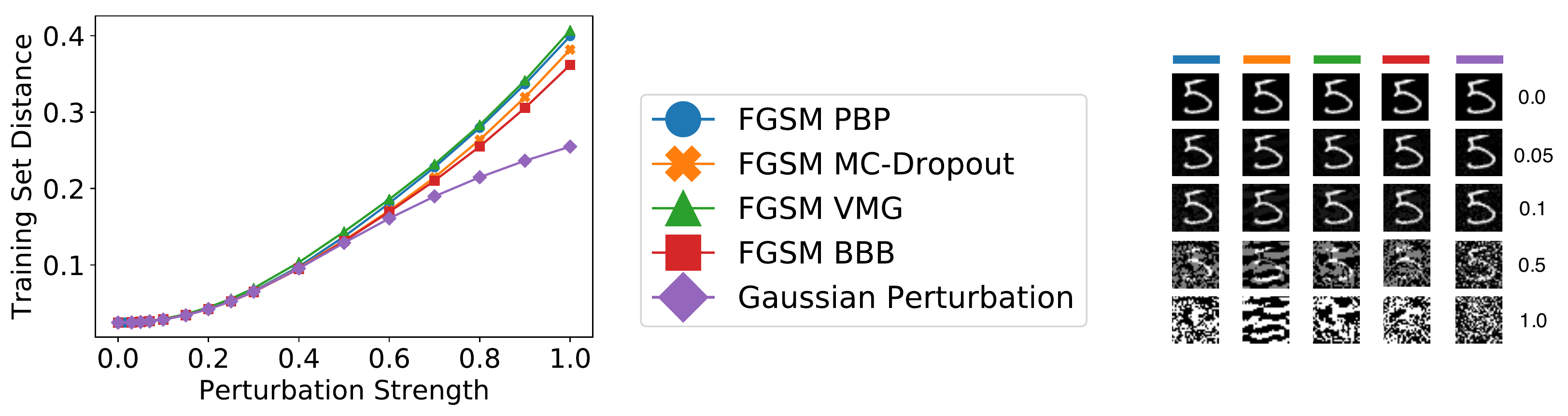}
    \caption{Mean training set distance for different sets generated with adversarial and Gaussian perturbations along with representative samples from each of the set for different perturbation strengths.}
    \label{fig:tsd}
\end{figure}

In order to compare Gaussian perturbation and adversarial perturbation on a common ground, we evaluate training set distance for different attack and noise strengths.
For a sample $x$, this distance is computed as average pixel-wise Euclidean distance of the nearest neighbour in the training set.
The mean distance across the sets is shown in Figure \ref{fig:tsd}.

\subsection{More Uncertainty Footprints}

\begin{figure}[!htb]
    \centering
    \includegraphics[width=\linewidth]{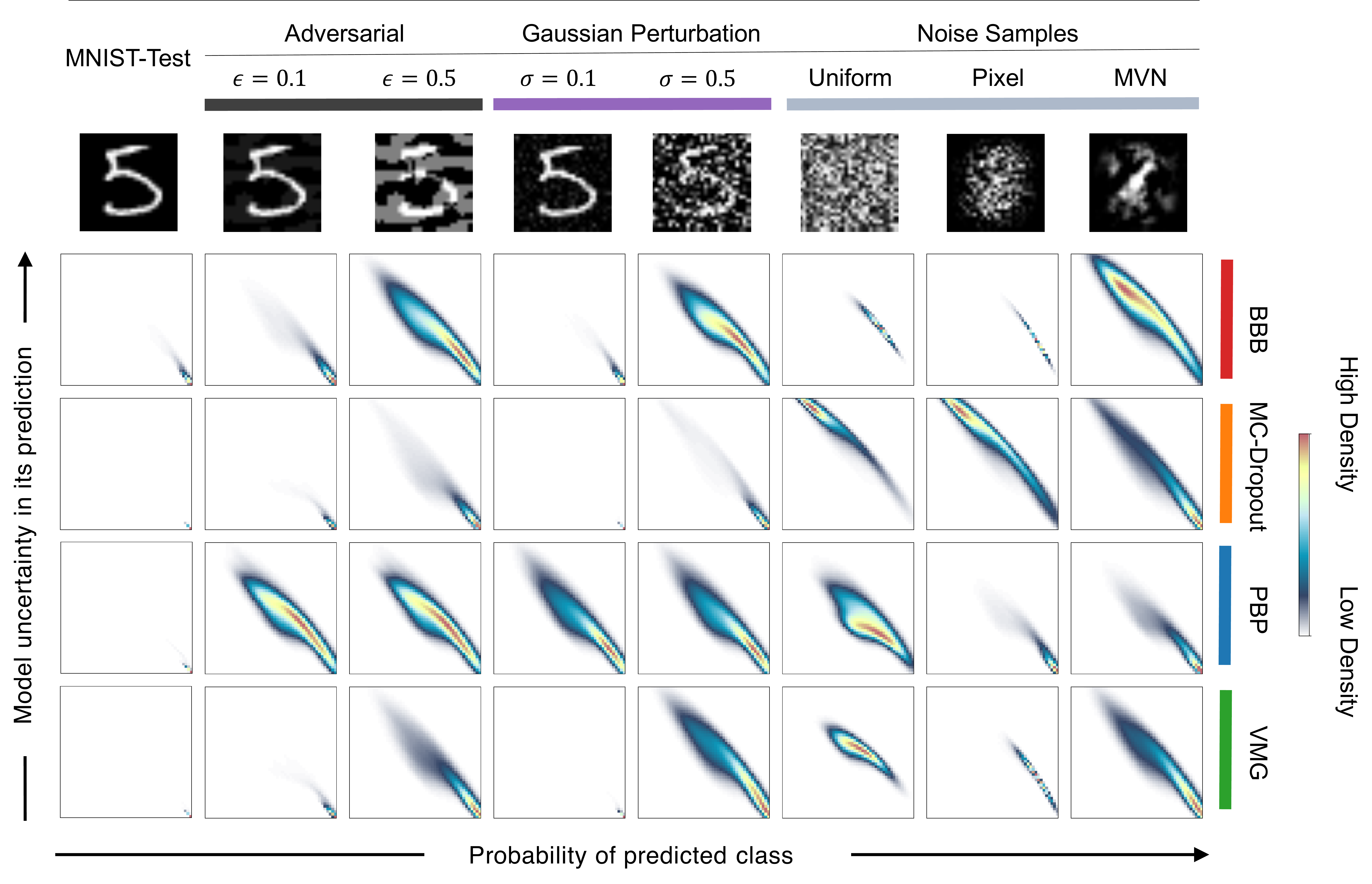}
    \caption{Scatter plot of dataset points along two summaries of uncertainty - i) probability of predicted class (x-axis) and predicted entropy (y-axis), overlayed with the density of points.
    A representative sample from each of the set is included for clarity.}
    \label{fig:footprint-etrpy}
\end{figure}

\begin{figure}[!htb]
    \centering
    \includegraphics[width=\linewidth]{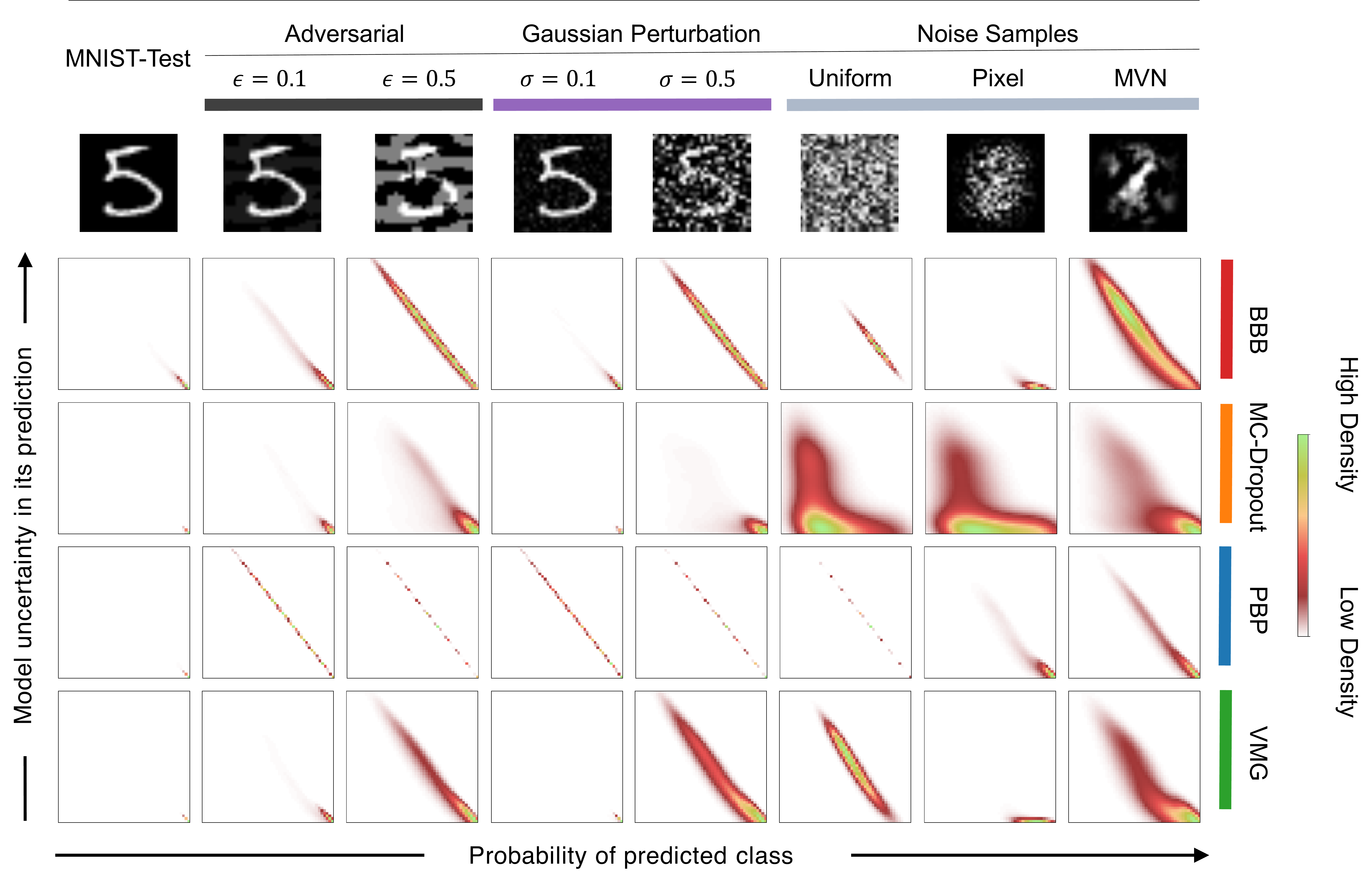}
    \caption{Scatter plot of dataset points along two summaries of uncertainty - i) probability of predicted class (x-axis) and variation ratio (y-axis), overlayed with the density of points.
    A representative sample from each of the set is included for clarity.}
    \label{fig:footprint-vr}
\end{figure}

Analogous to uncertainty footprints of MUMMI, we visualise the footprints for predicted entropy (Figure \ref{fig:footprint-etrpy}) and variation ratio (Figure \ref{fig:footprint-vr}).

\end{document}